\documentclass[letterpaper]{article}
\usepackage{uai2020}
\usepackage[margin=1in]{geometry}

\usepackage{times}

\usepackage[round]{natbib}

\usepackage[ruled]{algorithm2e}
\usepackage{amsfonts}       
\usepackage{nicefrac}       
\usepackage{microtype}      

\usepackage{arydshln}
\usepackage{booktabs}

\usepackage{amsmath}
\usepackage{amssymb}
\usepackage{subfig}
\usepackage{graphicx}
\usepackage{multirow}
\usepackage{cite}
\usepackage{color,soul}

\newcommand{\R}{\mathbb{R}}
\newcommand{\E}{\mathbb{E}}
\newcommand{\Lc}{\mathcal{L}}
\newcommand{\X}{\boldsymbol{x}}
\newcommand{\Z}{\boldsymbol{z}}
\newcommand{\T}{\boldsymbol{\theta}}
\newcommand{\PH}{\boldsymbol{\phi}}
\newcommand{\PS}{\boldsymbol{\psi}}
\newcommand{\ET}{\boldsymbol{\eta}}

\newcommand{\given}{\,|\,}

\newcommand{\squishlist}{
 \begin{list}{$\bullet$}
  { \setlength{\itemsep}{0pt}
     \setlength{\parsep}{1pt}
     \setlength{\topsep}{1pt}
     \setlength{\partopsep}{0pt}
     \setlength{\leftmargin}{1em}
     \setlength{\labelwidth}{0.5em}
     \setlength{\labelsep}{0.5em} } }

\newcommand{\squishend}{
  \end{list}  }

\usepackage[normalem]{ulem}

\title{Pairwise Supervised Hashing with 
Bernoulli Variational Auto-Encoder and Self-Control Gradient Estimator}

\author{\textbf{Siamak Zamani Dadaneh}$^{\dagger\star}$, \textbf{Shahin Boluki}$^{\dagger\star}$, \textbf{Mingzhang Yin}$^{\ddagger}$, \textbf{Mingyuan Zhou}$^{\ddagger}$, \textbf{Xiaoning Qian}$^{\dagger}$ \\
$^{\dagger}$ Texas A\&M University \quad $^{\ddagger}$The University of Texas at Austin \\  $^{\star}$Equal contribution}

\begin{document}

\maketitle

\begin{abstract}
  Semantic hashing has become a crucial component of fast similarity search in many large-scale information retrieval systems, in particular, for text data. Variational auto-encoders~(VAEs) with binary latent variables as hashing codes provide state-of-the-art performance in terms of precision for document retrieval. We propose a pairwise loss function with discrete latent VAE to reward within-class similarity and between-class dissimilarity for supervised hashing. Instead of solving the optimization relying on existing biased gradient estimators, an unbiased low-variance gradient estimator is adopted to optimize the hashing function by evaluating the non-differentiable loss function over two correlated sets of binary hashing codes to control the variance of gradient estimates. This new semantic hashing framework achieves superior performance compared to the state-of-the-arts, as demonstrated by our comprehensive experiments.  
\end{abstract}

\section{INTRODUCTION}
The problem of \emph{similarity search} is to find the most similar items in a large collection to a query item of interest \citep{andoni2009nearest}. Fast similarity search is at the core of many information retrieval applications, such as collaborative filtering \citep{sarwar2001item}, content-based retrieval \citep{lew2006content}, and caching \citep{pandey2009nearest}. In particular, with the explosion of information on Internet in the form of text data, searching for relevant content in such gigantic databases is critical. 

Traditional text similarity search methods are conducted in the space of original word counts, and thus can be computationally prohibitive due to high dimensions. Therefore, many research efforts have been devoted to employ approximate similarity search approaches in lower embedding dimensions. Semantic hashing \citep{salakhutdinov2009semantic} is an effective way of accelerating similarity search by designing compact binary codes in a low-dimensional space so that semantically similar documents are mapped to similar codes. The similarity between documents is evaluated by simply computing the pairwise Hamming distances between the hashing codes, i.e., the number of bits that are different between two codes. Furthermore, exploiting binary hashing codes is much more memory efficient, especially for big 
text corpora.

Deep learning has dramatically improved the state-of-the-arts in many applications, including speech recognition, computer vision,
and natural language processing \citep{lecun2015deep}. Learning expressive feature representations for complex data lies at the core of deep learning. Recently, deep generative models such as variational auto-encoder (VAE) have been proposed for neural semantic hashing \citep{chaidaroon2017variational}. Employing VAEs for document hashing has two major benefits. First, they can learn flexible nonlinear distributed representations of the original high-dimensional documents. Second, due to amortized computational cost for inference in VAEs, the hashing codes for new documents can be simply calculated with one pass through the encoder network.

In their basic form, VAEs assume that latent variables are distributed according to a multivariate normal distribution. The continuous latent representations are then binarized to obtain the hashing codes corresponding to the documents. As a result, the information contained in the continuous representations may be lost during the binarization step. \citet{shen2018nash} have developed a VAE framework with Bernoulli latent variables as hashing codes, obviating the need for the binarization step. To optimize the VAE model parameters, straight-through (ST) gradient estimator \citep{bengio2013estimating} with respect to binary latent variables is adopted in \citet{shen2018nash}. 
While easy to implement, ST gradient estimator is clearly biased, and hence it can undermine the performance of the VAE with binary latent representations as hashing codes to capture the semantic similarities of documents. 

In this paper, we aim to develop a faithful discrete VAE with Bernoulli latent variables as binary hashing codes that can be inferred without bias. When additional information such as document labels can be leveraged for a more targeted similarity search, we propose a pairwise supervised hashing (PSH) framework to derive better hashing codes, with two main objectives: (1) to learn informative binary codes, capable of reconstructing the original word counts; 
(2) to minimize the distance between the hashing codes of documents from the same class and maximize this distance for documents from different classes. The first objective can be achieved through maximizing the evidence lower bound (ELBO) with weighted Kullback--Leibler (KL) regularization \citep{alemi2018fixing,zhao2017infovae,higgins2017beta}. To achieve the second objective, we add a pairwise loss function to reward within-class similarity and between-class dissimilarity. This end-to-end generative framework is distinct from previous methods training a neural network classifier with latent variables as inputs and document labels as outputs for supervised hashing \citep{shen2018nash,chaidaroon2017variational}, which fail to extract useful similarity patterns for efficient search as they consider documents in isolation. 

We exploit stochastic gradient based optimization to learn this Bernoulli VAE hashing model. The main difficulty arises due to the binary hashing code based latent representations. The recently proposed augment-REINFORCE-merge (ARM) \citep{yin2019arm} gradient estimator provides a natural solution with unbiased low-variance gradient updates during the training of our discrete VAE. 
With a single Monte Carlo sample, the
estimated gradient is the product of uniform random noise and the difference of the objective functions with two vectors of correlated binary latent variables as inputs. Applying the ARM gradient leads to not only fast convergence, but also low negative evidence lower bounds for variational inference, thus increasing the ability to reconstruct the original word counts from the binary hashing codes.

Comprehensive experiments conducted on benchmark datasets for both supervised and unsupervised hashing demonstrate the superior performance of our proposed framework in terms of precision for document retrieval. In particular, PSH gains significantly better performance for short hashing codes making it more attractive for practical applications with limited memory budget.

Our main contributions to hashing-based similarity retrieval include:
\squishlist
  \item We propose a flexible discrete VAE-based framework, directly with binary hashing codes as latent representations, for both unsupervised and supervised semantic hashing. With unbiased and low-variance ARM gradient estimator, efficient variational inference as well as one-pass hashing code generation given new documents can be achieved without commonly adopted continuous relaxation. 
  \item A novel pairwise loss function is defined for supervised hashing, obviating the need for access to ordinal labels in the training phase. 
  ARM gradient estimator is specially useful for learning with such a loss function based on the expectation of non-differentiable functions with binary random variables for hashing codes. 
  \item Our method is highly scalable, applicable to large-scale data. Our comprehensive experimental results with ablation studies have verified the advantage of our direct hashing code based VAE with ARM variational inference, as well as the benefits from our new loss function with the expected pairwise loss. More importantly, our new method consistently outperforms state-of-the-art methods over several widely used benchmark datasets.  
\squishend  

The remainder of this paper is organized as follows. In Section~2, we present the main methodology, including the structure of Bernoulli VAE for document hashing, optimization using ARM gradient estimator, and pairwise hashing in the supervised scenario. Section~3 discusses related work. Section~4 provides comprehensive experimental results in supervised as well as unsupervised settings, with comparison with existing hashing methods. Section~5 concludes the paper.

\section{METHODS}
\label{sec:method}

\subsection{Hashing Using Bernoulli VAEs}
Let $\X$ and $\Z$ denote the input document and its corresponding binary hashing code. Specifically, $\X \in \mathcal{Z}_+^{|V|}$ is a vector of word counts for the input document, where $|V|$ is the size of the vocabulary $V$. Under the variational auto-encoder (VAE) framework  \citep{kingma2013auto,rezende2014stochastic}, a generative (decoding) model $p_{\T}(\X\given\Z)$ reconstructs the input document from the binary hashing code, while an inference (encoder) model $q_{\PH}(\Z\given\X)$ infers the code $\Z$ from the input document $\X$. The model parameters $\{\T, \PH\}$ are the weights of neural networks employed by the decoder and encoder.

\subsubsection{Decoder Structure}
To build the decoder, we follow the same procedure as in \citet{chaidaroon2017variational,shen2018nash}, and utilize a \emph{softmax} decoding function. Assuming that $t_i$, the $i$th token within document $\X$, is the $w$th word of the vocabulary, we denote its one-hot vector representation by $\boldsymbol{o}_w \in \{0,1\}^{|V|}$, a vector with a one at $w$th element and zeros elsewhere. The decoder network comprises
a linear transformation of the latent binary hashing code $\Z$, followed by a softmax function which outputs the likelihood of individual tokens as:
\begin{equation}
    p_{\T}(t_i=w\given\Z) = \frac{\exp \big( \Z^T E \boldsymbol{o}_w + b_w \big)}{\sum_{j=1}^{|V|} \exp \big( \Z^T E \boldsymbol{o}_j + b_j \big)}, \quad \text{for } w \in V,
    \label{eq:decoder}
\end{equation}
where $E \in \R^{K \times |V|}$ can be interpreted as a word embedding matrix and $\boldsymbol{b}=[b_1,...,b_{|V|}]^T$ are the word biases. Thus, the decoder parameters to be learned are $\T=\{E,\boldsymbol{b}\}$. Given the individual token likelihoods in (\ref{eq:decoder}) and the word counts $\X$, the document likelihood can be computed as
\begin{eqnarray}
    \log p_{\T}(\X\given\Z) &=& \sum_i \log p_{\T}(t_i\given\Z) \nonumber\\
    &=& \sum_{w: x_w>0} x_w \log p_{\T}(x_w\given\Z).
    \label{eq:dec}
\end{eqnarray}
To exploit the relevance of words in documents, we replace the log weights in (\ref{eq:dec}) with Term Frequency Inverse Document Frequency
(TF-IDF) \citep{ramos2003using}. Hence, we use the following modified reconstruction term in the optimization procedure of the ELBO explained in latter sections:
$$\E_{q_{\PH}(\Z\given\X)} \Big[ \sum_{w: x_w>0} \text{TF-IDF}_w \times \log p_{\T}(x_w\given\Z) \Big].$$ 


\subsubsection{Encoder Structure}
We employ the amortized inference of hashing codes for documents by constructing an inference network as $f_{\PH}(\X)$ to approximate the true posterior distribution $p(\Z\given\X)$ by $q_{\PH}(\Z\given\X)$. More precisely, the approximate posterior for the $K$-dimensional latent code $\Z \in \{0,1\}^K$ is expressed~as
\begin{equation}
    q_{\PH}(\Z\given\X) = \prod_{k=1}^K \mbox{Bernoulli}\big(z_k;\sigma(f_{\PH}(\X)_k)\big),\label{eq:Q}
\end{equation}
where $\sigma(\cdot)$ is the sigmoid function, and $f_{\PH}(\X)_k$ is the $k$th element of the encoder neural network's output. In the training phase, latent codes are sampled using the Bernoulli distributions in \eqref{eq:Q}  and subsequently fed into the decoder network, while in the testing phase, hard thresholding the means at 0.5 is used to infer the hashing codes. Finally, we place independent Bernoulli priors on the components of latent codes as $p(\Z) = \prod_{k=1}^K \mbox{Bernoulli}(z_k;\gamma_k)$, where $\gamma_k \in [0,1]$. Our Bernoulli distributed latent variables obviate the need for a separate binarization step; and hence they are more capable of capturing the semantic structure of input documents.

\subsubsection{Variational Inference}
To estimate the parameters of encoder and decoder networks, the VAE framework optimizes ELBO defined as:
\begin{eqnarray}
    \Lc(\T,\PH) &:=& \E_{p_{\mathcal{D}}(\X)} \Big[ \E_{q_{\PH}(\Z\given\X)} \big[ \log p_{\T}(\X\given\Z) \big] \nonumber\\
    & & - \mbox{KL}(q_{\PH}(\Z\given\X) || p(\Z)) \Big] \nonumber\\
    &\leq& \E_{p_{\mathcal{D}}(\X)} \big[  \log p_{\T}(\X) \big],
    \label{eq:elbo}
\end{eqnarray}
where KL is the Kullback--Leibler divergence and $p_{\mathcal{D}}(\X)$ is the empirical distribution of the inputs. Since both prior and approximate posterior are Bernoulli distributions, the KL term can be computed in the closed form:
\begin{align}
    &\mbox{KL}(q_{\PH}(\Z\given\X) || p(\Z)) = \sum_k \Big\{ \sigma(f_{\PH}(\X)_k) \log \frac{\sigma(f_{\PH}(\X)_k)}{\gamma_k} \nonumber\\&+ (1-\sigma(f_{\PH}(\X)_k)) \log \frac{1 - \sigma(f_{\PH}(\X)_k)}{1 - \gamma_k}  \Big\}.
    \label{eq:kl}
\end{align}

In practice, to extract useful latent representations and to avoid latent variable collapse \citep{dieng2019avoiding}, a modification of ELBO with the  weighted KL term is employed:
\begin{eqnarray}
    \Lc_{\lambda}(\T,\PH) &:=& \E_{p_{\mathcal{D}}(\X)} \Big[ \E_{q_{\PH}(\Z\given\X)} \big[ \log p_{\T}(\X\given\Z) \big] \nonumber\\
    & & - \lambda \mbox{KL}(q_{\PH}(\Z\given\X) || p(\Z)) \Big] \nonumber,
\end{eqnarray}
where $0 < \lambda < 1$. The parameters are then estimated by stochastic gradient optimization of $\Lc_{\lambda}(\T,\PH)$. In what follows, we drop the expectation with respect to the empirical distribution to simplify the notations.

\subsection{Pairwise Supervised Hashing (PSH)}
When training data come with side information such as document labels, the previously discussed discrete VAE is not ready to take advantage of that. To mitigate such a shortcoming for deriving better latent hashing codes in this generative framework, we add a supervised layer: Let $y$ denote the label for the input document $\X$. Given a neural network $f_{\ET}$ parameterized by $\ET$, which takes as input the latent hashing code $\Z$ and predicts the document label, the supervised hashing objective to be minimized can be expressed as:
\begin{equation}
    -\Lc_{\lambda}(\T,\PH) + \alpha \E_{q_{\PH}(\Z\given\X)} \big[ \Lc'(y;f_{\ET}(\Z))\big],
\end{equation}
where $\alpha > 0$ is a hyperparameter and $\Lc'(y;f_{\ET}(\Z))$ is the cross entropy loss function for label prediction.

\begin{figure}[t]
    \centering
    \resizebox{\columnwidth}{!}{%
    \includegraphics{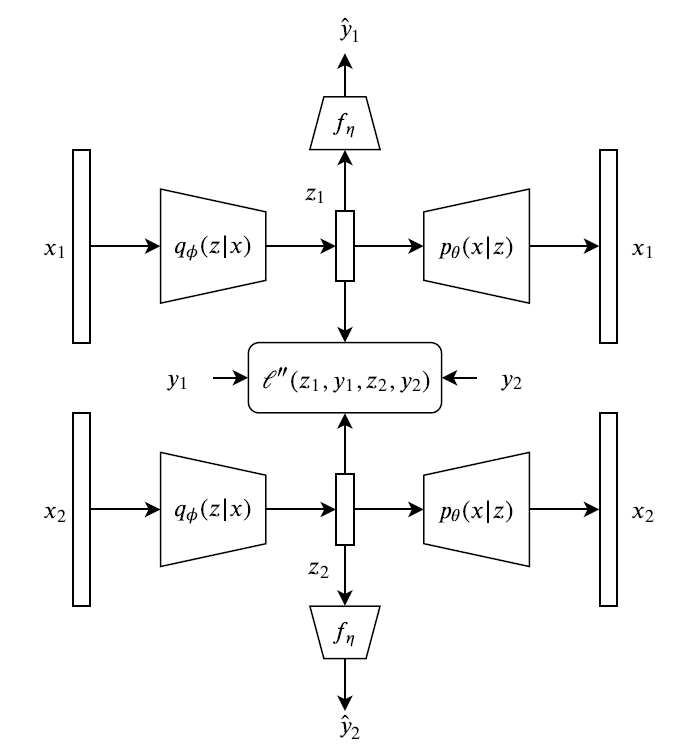}
    }%
    \caption{The graphical representation of Pairwise Supervised Hashing (PSH) model. Documents $\X_1$ and $\X_2$ each go through the encoder network ($q_{\PH}(\Z\given\X)$) to generate latent hashing codes $\Z_1$ and $\Z_2$, respectively. Each hashing code then goes through decoder ($p_{\T}(\X\given\Z)$) and classifier ($f_{\ET}$) networks to reconstruct the input document and predict its label, respectively.}
    \label{fig:psh}
\end{figure}

To further improve the performance of supervised hashing, we propose a pairwise supervised hashing (PSH) training framework. The core idea of PSH is to minimize the distance between latent codes of similar documents and simultaneously maximize the distance between latent codes of documents which fall into different categories. Denoting $(\X^{(1)},y^{(1)})$ and $(\X^{(2)},y^{(2)})$ as two randomly sampled documents with their corresponding latent codes $\Z^{(1)}$ and $\Z^{(2)}$, PSH places an extra loss function as:
\begin{eqnarray}
    \Lc''(\Z^{(1)},\Z^{(2)}) &=& \mathbf{1}_{y^{(1)}=y^{(2)}}d(\Z^{(1)},\Z^{(2)}) \nonumber\\
    && -\mathbf{1}_{y^{(1)} \neq y^{(2)}}d(\Z^{(1)},\Z^{(2)}),
\end{eqnarray}
where $d(\cdot,\cdot)$ is a distance metric and $\mathbf{1}_{S}$ is the indicator function being equal to one when $S$ is true. The final objective function for the PSH is thus
\begin{eqnarray}
    \Lc_{PSH}(\T,\PH) =& -&\big[ \Lc^{(1)}_{\lambda}(\T,\PH) + \Lc^{(2)}_{\lambda}(\T,\PH) \big] \nonumber\\
    &+& \E_{\prod_{t=1}^2q_{\PH}(\Z^{(t)}\given\X^{(t)})}\Big[ \beta \Lc''(\Z^{(1)},\Z^{(2)}) \nonumber\\
    &+& \alpha \big[\Lc'(y^{(1)};f_{\ET}(\Z^{(1)})) \nonumber\\
    &+& \Lc'(y^{(2)};f_{\ET}(\Z^{(2)}))\big] \Big],
    \label{eq:sph}
\end{eqnarray}
where $\Lc^{(t)}_{\lambda}(\T,\PH)$ is the ELBO for document $t$ and $\beta > 0$ is a hyperparameter. In practice, effective hyperaparameters for PSH can be determined by cross validation. The graphical representation of PSH is shown in Figure~\ref{fig:psh}.

\subsection{Gradient Updates for Training}
Optimizing the PSH loss function (\ref{eq:sph}) is difficult, as the backpropagation algorithm cannot be applied to the discrete Bernoulli sampling layers. In this section, we first present two widely used gradient estimators for discrete latent variables. Then, we present how ARM, an unbiased gradient estimator, can be employed for backpropagation through discrete layers of our PSH framework.

\subsubsection{Straight-Through Gradient Estimator}
The straight-through (ST) gradient estimator \citep{bengio2013estimating} simply backpropagates through a discrete sampling unit as if  had been the identity function. More precisely, given the input document $\X$, first the binary latent representation is  sampled as
$$\Z \sim \mbox{Bernoulli}(\sigma(f_{\PH}(\X))),$$
and then the input to the decoder is calculated as
$$\Z' = \mbox{Stop Gradient} \Big( \Z - \sigma(f_{\PH}(\X)) \Big) + \sigma(f_{\PH}(\X)),$$
where the terms inside the \emph{Stop Gradient} operator are considered as constants in the backpropagation step \citep{bengio2013estimating}.

Although this is clearly a biased estimator, it is simple to implement and fast, with good performance in practice. 


\subsubsection{Gumbel-Softmax Gradient Estimator}
The Gumbel-Softmax (GS) distribution \citep{jang2016categorical,maddison2016concrete}, a continuous distribution on the simplex, can be adopted to approximate the gradient estimates of the loss functions involving categorical samples, where parameter gradients can be computed via the reparameterization trick \citep{kingma2013auto}. Consider an inference network architecture that for each component of latent hashing code $z_k$, it outputs the ratio of the probabilities of being one or zero as $\pi_{k}=\frac{\pi_{k1}}{\pi_{k0}}$. The binary representation of $z_k$ can be obtained using the Gumbel-Max trick and the fact that the difference of two Gumbels is a Logistic distribution:
$$z_k = \text{unit-step}(g + \log \pi_{k}),$$
where $g$ is a sample drawn from Logistic which can be generated as $g = \log u - \log (1-u)$ with $u \sim \text{Uniform}(u;0,1)$. In the backward pass of backpropagation, the binary random variables are replaced with continuous, differentiable variables as:
\begin{equation}
    h_{k} = \frac{1}{1+\exp(-(g + \log \pi_{k})/\tau)},
\end{equation}
where $\tau>0$ is the \emph{temperature}. As the softmax temperature $\tau$ approaches zero, samples from the Gumbel-Softmax distribution become one-hot and the Gumbel-Softmax distribution becomes identical to the Bernoulli distribution.

\subsubsection{Self-Control Gradient Estimator with ARM}
Both ST and GS approximations lead to biased gradient estimates. To reliably derive latent codes in our PSH framework by backpropagating unbiased gradients through stochastic binary units, we employ the ARM estimator that is unbiased, exhibits low variance, and has low computational complexity  \citep{yin2019arm,boluki2020learnable,icassp_arsm}. More importantly, unlike ST and GS gradient estimators, it can be applied to non-differentiable objective functions, tailored to training discrete VAEs with the PSH loss function $\Lc_{PSH}$.

Given a vector of $K$ binary random variables $\Z = (z_1,...,z_K)^T$, the gradient of the objective function
$$\mathcal{E}(\PS) = \mathbb{E}_{\Z \sim \prod_{k=1}^ K \text{Bernoulli}(z_k;\sigma(\psi_k))} \big[ f(\Z) \big]$$
with respect to $\PS=(\psi_1,...,\psi_K)^T$, the logits of the Bernoulli probability parameters, can be expressed as
\begin{eqnarray}
    \nabla_{\PS} \mathcal{E}(\PS) &=& \mathbb{E}_{\boldsymbol{u} \sim \prod_{k=1}^ K \text{Uniform}(u_k;0,1)} \Big[ \nonumber\\
    && \Big( f(\boldsymbol{1}_{\boldsymbol{u}>\sigma(-\PS)}) - f(\boldsymbol{1}_{\boldsymbol{u}<\sigma(\PS)})\Big) \nonumber\\
    &\times& (\boldsymbol{u} - 1/2) \Big ],
    \label{eq:arm}
\end{eqnarray}
where $\boldsymbol{1}_{\boldsymbol{u}>\sigma(-\PS)} := \big(\boldsymbol{1}_{u_1>\sigma(-\psi_1)},...,\boldsymbol{1}_{u>\sigma(-\psi_K)}  \big)^T$, and the function $f(\cdot)$ does not need to be differentiable. Note that $\boldsymbol{1}_{\boldsymbol{u}>\sigma(-\PS)}$ and  $\boldsymbol{1}_{\boldsymbol{u}<\sigma(\PS)}$ are two correlated binary vectors, which are evaluated under $f(\cdot)$ and then used to control the gradient variance. Thus we can consider ARM  as a self-control gradient estimator that does not need extra baselines with learnable parameters for variance reduction.

The training steps of PSH with ARM gradient estimator are presented in Algorithm~1. It starts with sampling two mini-batches of input documents with the same size, randomly. The documents then go through the encoder network to obtain the Bernoulli logits, and the binary latent hashing codes are generated using the Bernoulli distribution. For documents in each mini-batch, the gradients of the reconstruction and KL terms with respect to the parameters of the encoder network are calculated using the ARM   estimator in \eqref{eq:arm} with a single Monte Carlo sample and the closed form in (\ref{eq:kl}), respectively. Lastly, the gradient of the pairwise loss term is also calculated using the ARM estimator in \eqref{eq:arm} with a single Monte Carlo sample. These gradients are combined to update the parameters at each iteration.

\begin{algorithm}[t]
\caption{Pairwise Supervised Hashing with ARM gradient estimator.}
\SetAlgoLined
\KwIn{Data $\{\X\}$, neural networks $f_{\PH}$ (encoder), $f_{\T}$ (decoder) and $f_{\ET}$ (classifier), step size $\rho$.}
\KwOut{Model parameters $\PH$, $\T$ and $\ET$.}
Initialize model parameters randomly. \\
\While{not converged}{
Sample two mini-batches of data. \\
\For{each mini-batch}{
$\PS = f_{\PH}(\X)$ \\
Sample $\Z \sim \mbox{Bernoulli}(\PS)$. \\
Calculate $\nabla_{\PH}\mbox{KL}$, the gradient of $\mbox{KL}$ in~(\ref{eq:kl}). \\
Calculate $\nabla_{\PS}\Lc_{\lambda}^{(r)}$, the gradient of reconstruction term ($\Lc_{\lambda}^{(r)}$) in~(\ref{eq:elbo}) using ARM (\ref{eq:arm}).\\
$\nabla_{\PH}\Lc_{\lambda}^{(r)} = \sum_{k} (\nabla_{\psi_k}\Lc_{\lambda}^{(r)}) (\nabla_{\PH} \psi_k)$ \\
$\nabla_{\PH}\Lc_{\lambda} = \nabla_{\PH}\Lc_{\lambda}^{(r)} + \lambda \nabla_{\PH}\mbox{KL}$ \\
Calculate $\nabla_{\T}\Lc_{\lambda}^{(r)}$ and $\nabla_{\ET}\mathcal{L}'$\\
}
Calculate the pairwise loss gradients $\nabla_{\PH,\T}\mathcal{L}''$\\
Combine the gradients to form $\nabla_{\PH,\T,\ET} \mathcal{L}_{PSH}$\\
$\PH = \PH + \rho \nabla_{\PH} \mathcal{L}_{PSH} $ \\
$\T = \T + \rho \nabla_{\T} \mathcal{L}_{PSH} $ \\
$\ET = \ET + \rho \nabla_{\ET} \mathcal{L}_{PSH} $
}
\end{algorithm}

\section{RELATED WORK}
Current hashing methods can be categorized into two groups; data-dependent and data-independent. Locally sensitive hashing (LSH) \citep{datar2004locality} is a data-independent hashing method, with asymptotic theoretical properties leading to performance guarantees. LSH, however, usually requires long hashing codes to achieve satisfactory performance. To achieve more effective hashing codes, recently data-dependent machine learning methods are proposed, ranging from unsupervised and supervised to semi-supervised settings.

Unsupervised hashing methods such as Spectral
Hashing (SpH) \citep{weiss2009spectral}, graph hashing \citep{liu2011hashing},  and self taught hashing (STH) \citep{zhang2010self} attempt to extract the data properties, such as distributions and latent manifold structures to design compact codes with improved precision. Supervised hashing methods such as semantic hashing using tags and topic modeling (SHTTM) \citep{wang2013semantic} and kernel-based supervised hashing (KSH) \citep{liu2012supervised} attempt to leverage label/tag information for hashing function learning. A semi-supervised learning approach was also employed to design hashing functions by exploiting both labeled and unlabeled data \citep{wang2010semi}.

Recently, deep learning based methods have gained attraction for the hashing problem. Variational deep semantic hashing (VDSH) \citep{chaidaroon2017variational} uses a VAE to learn the latent representations of documents and then uses a separate step to cast the continuous representations into binary codes. While fairly successful, this generative hashing model requires a two-stage training. Neural architecture for semantic hashing (NASH) \citep{shen2018nash} proposed to substitute the Gaussian prior in VDSH with a Bernoulli prior to tackle this problem, by using a straight-through estimator \citep{bengio2013estimating} to estimate the gradient of neural network involving the binary variables. 

In this work, we exploit ARM~\citep{yin2019arm} gradient estimator to obtain unbiased low-variance gradient updates during the training of our discrete VAE. We further propose a pairwise loss function with the discrete latent VAE to reward within-class similarity and between-class dissimilarity for supervised hashing.

\section{EXPERIMENTAL RESULTS}
\label{sec:results}

\subsection{Datasets and Baselines}
We use three public benchmarks to evaluate the performance of our PSH and compare with other state-of-the-arts: \emph{Reuters21578} and \emph{20Newsgroups}, which are collections of news documents, as well as \emph{TMC} from SIAM text mining competition, containing air traffic reports provided by NASA. Properties of these datasets are included in Table~\ref{tab:data}. To make a direct comparison with
existing methods, we have employed the TFIDF features on these datasets.

We evaluate the performance of our discrete latent VAEs on both unsupervised and supervised semantic hashing tasks. We consider the following unsupervised baselines for comparison: locality sensitive hashing~(LSH)~\citep{datar2004locality}, stack restricted Boltzmann machines (S-RBM) \citep{salakhutdinov2009semantic}, spectral hashing (SpH) \citep{weiss2009spectral}, self-taught hashing (STH) \citep{zhang2010self}, variational deep semantic hashing (VDSH) \citep{chaidaroon2017variational}, and neural architecture for  semantic hashing~(NASH)~\citep{shen2018nash}.

For supervised semantic hashing, we compare the performance of PSH against a number of baselines: Supervised Hashing with Kernels (KSH) \citep{liu2012supervised}, Semantic Hashing using Tags and Topic Modeling (SHTTM) \citep{wang2013semantic}, Supervised Variational Deep Semantic Hashing (VDSH-S) \citep{chaidaroon2017variational}, VDSH-S with document-specific latent variable (VDSH-SP) \citep{chaidaroon2017variational}, and Supervised Neural Architecture for Semantic Hashing (NASH-DN-S) \citep{shen2018nash}.

\begin{table}[t]
    \centering
    \caption{Properties of three datasets in the experiments.}\vspace{-0.05in}
    \resizebox{\columnwidth}{!}{%
    \begin{tabular}{c||c|c|c}
    \hline
         Dataset & \#documents & vocabulary size & \#categories \\
         \hline\hline
         Reuters21578 & 10,788 & 10,000 & 20 \\
         20Newsgroups  & 18,828 & 7,164 & 20 \\
         TMC  & 21,519 & 20,000 & 22 \\
    \hline
    \end{tabular}
    }
    \label{tab:data}
\end{table}

\subsection{Implementation Details}
For the encoder networks, we employ a fully connected neural network with two hidden layers, both with 500 units and the ReLU nonlinear activation function. We train PSH using the Adam optimizer \citep{kingma2014adam} with a learning rate of $5 \times 10^{-4}$. Dropout \citep{srivastava2014dropout} is employed on the output
of encoder networks, with the dropping rate of 0.2. To facilitate comparisons with previous methods, we set the hashing code length to 8, 16, 32, 64, or 128, respectively. For all datasets, we use a KL weight of $\lambda=0.01$ for PSH, set the hyperparameters as $\beta=5 \times 10^{-2}$, and start with $\alpha=0.01$ and gradually increase its value to 0.1. The temperature of Gumbel-Softmax gradient estimator is initialized with 1, and it is gradually decreased with a decay rate of 0.96, until it reaches the minimum value of 0.1.

\subsection{Evaluation Metric}
To evaluate the quality of hashing codes for similarity
search, we follow previous works \citep{shen2018nash,chaidaroon2017variational} and consider each document in the test set as a query document. Specifically, the performance of different methods are measured with the \emph{precision at 100} metric as explained in the following. In the testing phase, we first retrieve the 100 nearest documents to the query document according to the Hamming distances of their corresponding hashing codes. We then calculate the percentage of documents among the 100 retrieved ones that belong to the same label (topic) with the query document. The ratio of the number of relevant documents to the number of retrieved documents is calculated as the precision score. The precision scores are further averaged over all test (query) documents.

\subsection{Results and Discussions}

\subsubsection{Unsupervised Hashing}

To examine how our discrete latent VAE with the ARM gradient estimator affects the quality of hashing codes, we evaluate its performance in an unsupervised scenario. More specifically, we build a binary VAE with the weighted KL regularization term on the training documents, and then use the trained encoder network to generate the binary hashing codes. To improve the performance of unsupervised hashing with VAE, we follow the procedure in \citet{shen2018nash}, and add a data-dependent noise to the binary hashing code before feeding it into the decoder network. 

Tables~\ref{tab:reu}, \ref{tab:20ng}, and \ref{tab:tmc} show the performance of the proposed ARM-facilitated discrete latent VAE (hereby referred to as ARM-DVAE) and baseline models on Reuters, 20 Newsgroup and TMC datasets respectively, under the unsupervised setting, with the number of hashing bits ranging from 8 to 128. It can be observed that exploiting the unbiased and low-variance ARM gradient estimator improves the performance of unsupervised hashing in terms of the retrieval precision in the majority of cases for these datasets. In particular, for the 128-bit hashing codes, ARM-DVAE improves the performance of NASH 22\% across all datasets, on average. These observations strongly support the remarkable benefit of using ARM gradient estimator to learn useful semantic hashing codes in the discrete latent VAE framework.

\begin{table}[t]
    \centering
    \caption{The performances of different unsupervised hashing models on the  Reuters dataset with different lengths of hashing codes.}\vspace{-0.05in}
    \resizebox{\columnwidth}{!}{%
    \begin{tabular}{c||c|c|c|c|c}
    \hline
         Method & 8 bits & 16 bits & 32 bits & 64 bits & 128 bits  \\
         \hline\hline
         LSH & 0.2802 & 0.3215 & 0.3862 & 0.4667 & 0.5194 \\
         S-RBM & 0.5113 & 0.5740 & 0.6154 & 0.6177 & 0.6452 \\
         SpH & 0.6080 & 0.6340 & 0.6513 & 0.6290 & 0.6045 \\
         STH & 0.6616 & 0.7351 & 0.7554 & 0.7350 & 0.6986 \\
         VDSH & 0.6859 & 0.7165 & 0.7753 & 0.7456 & 0.7318 \\
         NASH & \textbf{0.7113}  & \textbf{0.7624} &  0.7993 & 0.7812  & 0.7559 \\ 
         \hline
         ARM-DVAE & 0.6549 & 0.7455  & \textbf{0.8086}  & \textbf{0.8237} & \textbf{0.8230} \\
         \hline\hline
    
    \hline
    \end{tabular}%
    }
    \label{tab:reu}
\end{table}

\begin{table}[h]
    \centering
    \caption{The performances of different unsupervised hashing models on the 20 Newsgroup dataset with different lengths of hashing codes.}\vspace{-0.05in}
    \resizebox{\columnwidth}{!}{%
    \begin{tabular}{c||c|c|c|c|c}
    \hline
         Method & 8 bits & 16 bits & 32 bits & 64 bits & 128 bits  \\
         \hline\hline
         LSH &  0.0578 &  0.0597 &  0.0666  & 0.0770  & 0.0949 \\
         S-RBM  & 0.0594  & 0.0604 &  0.0533 &  0.0623 &  0.0642 \\
         SpH  & 0.2545 &  0.3200  & 0.3709 &  0.3196 &  0.2716 \\
         STH &  0.3664  & \textbf{0.5237} &  \textbf{0.5860} &  0.5806 &  0.5443 \\
         VDSH &  0.3643  & 0.3904  & 0.4327 &  0.1731 &  0.0522 \\
         NASH & 0.3786 & 0.5108 & 0.5671 & 0.5071 & 0.4664 \\ 
         \hline
         ARM-DVAE  & \textbf{0.3907} & 0.5074 & 0.5787 & \textbf{0.6224} & \textbf{0.6214} \\
         \hline\hline
         
    \hline
    \end{tabular}%
    }
    \label{tab:20ng}
\end{table}

\begin{table}[h]
    \centering
    \caption{The performances of different unsupervised hashing models on TMC dataset with different lengths of hashing codes.}\vspace{-0.05in}
    \resizebox{\columnwidth}{!}{%
    \begin{tabular}{c||c|c|c|c|c}
    \hline
         Method & 8 bits & 16 bits & 32 bits & 64 bits & 128 bits  \\
         \hline\hline
         LSH  & 0.4388  & 0.4393  & 0.4514 &  0.4553 &  0.4773 \\
         S-RBM &  0.4846 &  0.5108 &  0.5166 &  0.5190 &  0.5137 \\
         SpH &  0.5807 &  0.6055 &  0.6281 &  0.6143  & 0.5891 \\
         STH &  0.3723 &  0.3947 &  0.4105  & 0.4181 &  0.4123 \\
         VDSH &  0.4330 &  \textbf{0.6853} &  0.7108 &  0.4410  & 0.5847 \\
         NASH & 0.5849 & 0.6573 & 0.6921 & 0.6548 & 0.5998 \\ 
         \hline
         ARM-DVAE & \textbf{0.6239} & 0.6825 & \textbf{0.7362} & \textbf{0.7541} & \textbf{0.7599}\\
         \hline\hline
    
    \hline
    \end{tabular}%
    }
    \label{tab:tmc}
\end{table}

\begin{figure}[!th]
    \centering
    \subfloat{%
    \includegraphics[clip,width=0.77\columnwidth]{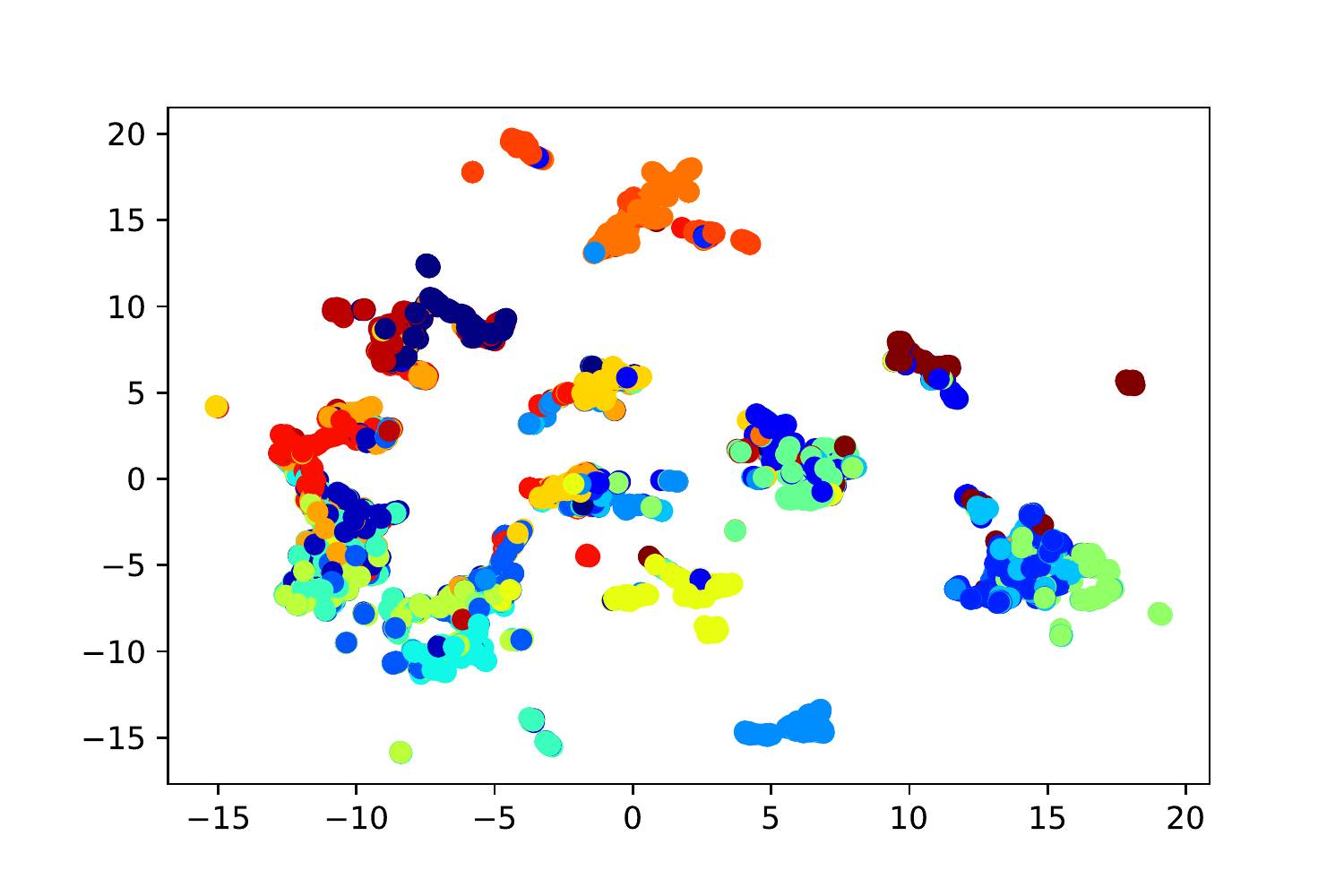}%
    }
    \qquad
    \subfloat{%
    \includegraphics[clip,width=0.77\columnwidth]{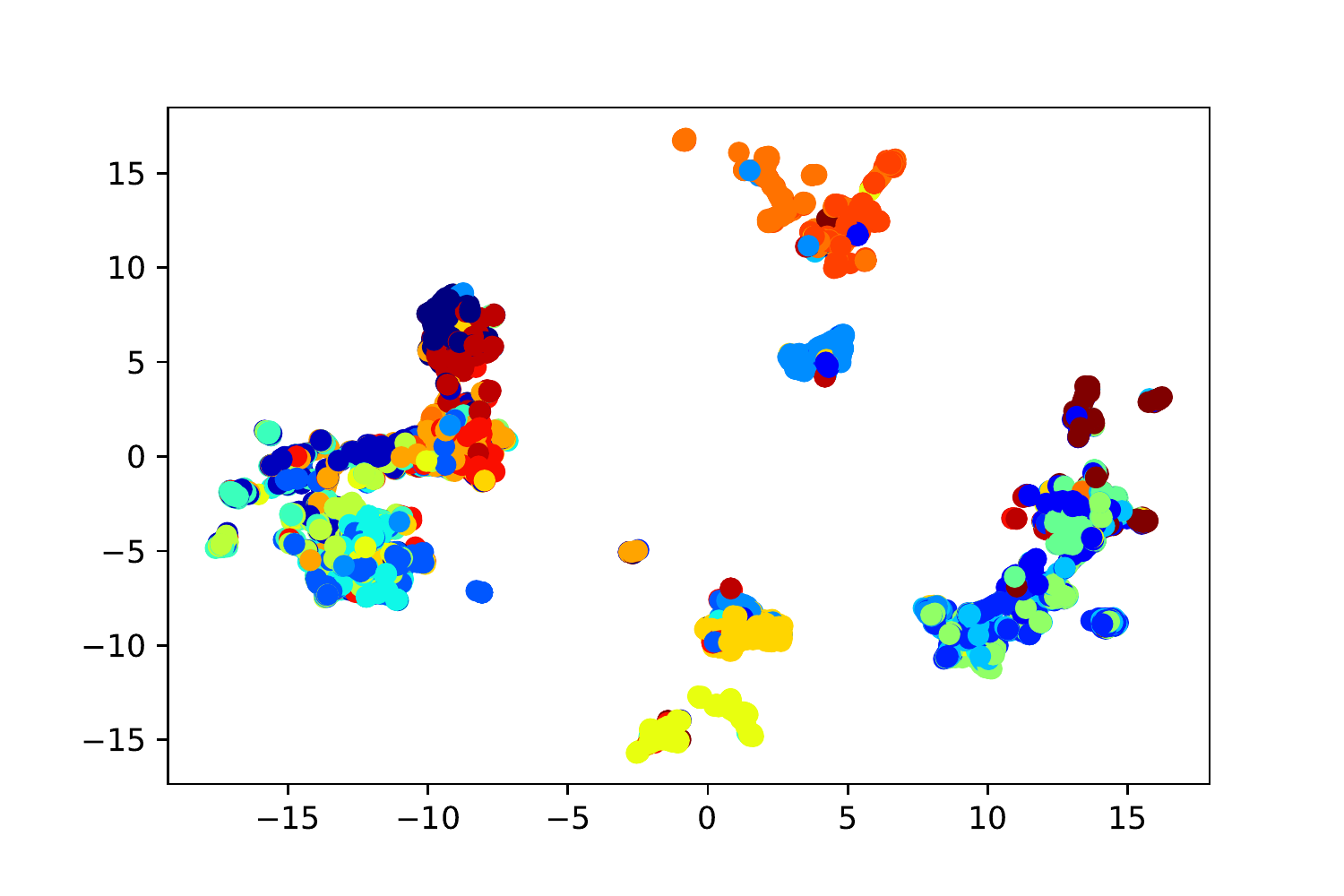}%
    }
    \qquad
    \subfloat{%
    \includegraphics[clip,width=0.77\columnwidth]{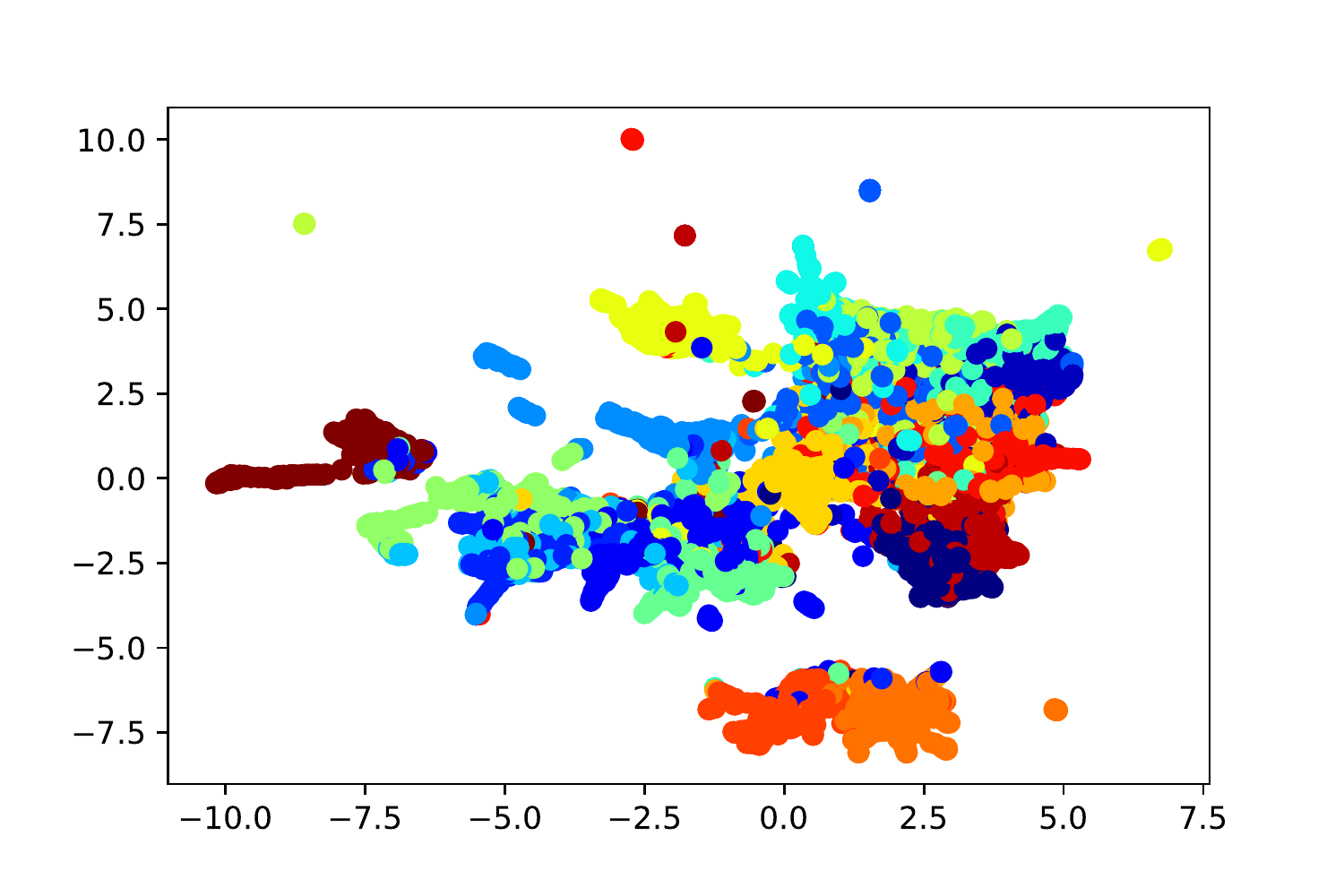}%
    }
    \caption{Visualization of the 32-dimensional latent semantic embeddings learned by ARM-DVAE (top), NASH (middle) and VDSH (bottom) for the 20 Newsgroups dataset. Each data point denotes a document, with each color representing one category. ARM-DVAE shows better separation of categories in the latent space. }
    \label{fig:vis}
\end{figure}

To further examine the performance of our ARM-facilitated discrete VAE in achieving effective document hashing, we illustrate the learned latent representations of ARM-DVAE, NASH and VDSH on the 20 Newsgroup dataset in Figure~\ref{fig:vis}. UMAP \citep{mcinnes2018umap} is used to project the 32-dimensional latent representations into a 2-dimensional space. In this figure, each data denotes a document, with each color representing one category. It can be observed that our ARM-DVAE is able to distinguish different categories of documents better than NASH with ST gradient estimator, and VDSH that binarizes normally distributed latent variables to obtain hashing codes. In particular, hashing codes from VDSH fail to form discernible clusters, confirming the advantage of using Bernoulli random variables for semantic hashing.

\subsubsection{Supervised Hashing}
Tables~\ref{tab:sreu}, \ref{tab:s20ng}, and \ref{tab:stmc} show the performance of the proposed and baseline models on the three datasets under the supervised setting, with the number of hashing bits ranging from 8 to 128. From these experimental results, it can be seen that for almost all datasets and hashing code lengths, the proposed PSH model outperforms all other methods in terms of retrieval precision. In particular, in 20 Newsgroup and TMC datasets, PSH with the ARM gradient estimator consistently outperforms other hashing methods by  large margins. This observation signifies the role of the ARM gradient estimator to obtain effective hashing functions. 

An interesting property of PSH, compared with its base discrete latent VAE models, is that it preserves the superior performance for both short and long hashing codes. For short hashing codes, this suggests the effectiveness of PSH, especially with the ARM gradient estimator, in learning useful and compact semantic latent representations of documents. For longer hashing codes, the performance of baseline methods tend to drop slightly. This phenomenon is attributed to the fact that for longer codes, the number of data points that are assigned to a certain binary code decreases exponentially. As a result, many queries may fail to return any neighbor documents \citep{shen2018nash}. The results here, however, indicate that PSH does not suffer from this phenomenon, suggesting the mitigating role of the pairwise loss term.

\begin{table}[t]
    \centering
    \caption{The performances of different supervised hashing models on Reuters dataset under different lengths of hashing codes.}\vspace{-0.05in}
    \resizebox{\columnwidth}{!}{%
    \begin{tabular}{c||c|c|c|c|c}
    \hline
         Method & 8 bits & 16 bits & 32 bits & 64 bits & 128 bits  \\
         \hline\hline
         KSH & 0.7840 & 0.8376 & 0.8480 & 0.8537 & 0.8620 \\
         SHTTM  & 0.7992 & 0.8520 & 0.8323 & 0.8271 & 0.8150 \\
         VDSH-S  & 0.9005 & 0.9121 & 0.9337 & 0.9407 & 0.9299 \\
         VDSH-SP  & 0.8890 & 0.9326 & 0.9283 & 0.9286 & 0.9395 \\
         NASH-DN-S  & 0.9214 & 0.9327 & 0.9380 & 0.9427 & 0.9336 \\
         \hline
         PSH-GS  & 0.8785 & \textbf{0.9604} & \textbf{0.9544} & \textbf{0.9594} & 0.9528 \\
         PSH-ARM & \textbf{0.9268}  & 0.9458  & 0.9451  &  0.9543 & \textbf{0.9569} \\
         \hline\hline
    
    \hline
    \end{tabular}%
    }
    \label{tab:sreu}
\end{table}

\begin{table}[h]
    \centering
    \caption{The performances of different supervised hashing models on 20 Newsgroup dataset under different lengths of hashing codes.}\vspace{-0.05in}
    \resizebox{\columnwidth}{!}{%
    \begin{tabular}{c||c|c|c|c|c}
    \hline
         Method & 8 bits & 16 bits & 32 bits & 64 bits & 128 bits  \\
         \hline\hline
         KSH & 0.4257 & 0.5559 & 0.6103 & 0.6488 & 0.6638 \\
         SHTTM & 0.2690 & 0.3235 & 0.2357 & 0.1411 & 0.1299 \\
         VDSH-S & 0.6586 & 0.6791 & 0.7564 & 0.6850 & 0.6916 \\
         VDSH-SP & 0.6609 & 0.6551 & 0.7125 & 0.7045 & 0.7117 \\
         NASH-DN-S & 0.6247 & 0.6973  & 0.8069 & 0.8213 & 0.7840 \\
         \hline
         PSH-GS  & 0.7387 & 0.8075 & 0.8274 & 0.8295 & 0.8271 \\
         PSH-ARM & \textbf{0.7507} & \textbf{0.8212}  & \textbf{0.8376}  & \textbf{0.8404}  & \textbf{0.8432} \\
         \hline\hline
    \end{tabular}%
    }
    \label{tab:s20ng}
\end{table}

\begin{table}[h]
    \centering
    \caption{The performances of different supervised hashing models on TMC dataset under different lengths of hashing codes.}\vspace{-0.05in}
    \resizebox{\columnwidth}{!}{%
    \begin{tabular}{c||c|c|c|c|c}
    \hline
         Method & 8 bits & 16 bits & 32 bits & 64 bits & 128 bits  \\
         \hline\hline
         KSH & 0.6608 & 0.6842 & 0.7047 & 0.7175 & 0.7243 \\
         SHTTM & 0.6299 & 0.6571 & 0.6485 & 0.6893 & 0.6474 \\
         VDSH-S & 0.7387 & 0.7887 & 0.7883 & 0.7967 & 0.8018 \\
         VDSH-SP & 0.7498 & 0.7798 & 0.7891 & 0.7888 & 0.7970 \\
         NASH-DN-S & 0.7438 & 0.7946 & 0.7987 & 0.8014 & 0.8139 \\
         \hline
         PSH-GS  & 0.7931 & 0.8189 & 0.8314 & 0.8379 & 0.8426 \\
         PSH-ARM &  \textbf{0.8010} & \textbf{0.8329}  & \textbf{0.8524}  & \textbf{0.8565}  & \textbf{0.8617} \\
         \hline\hline
    
    \hline
    \end{tabular}%
    }
    \label{tab:stmc}
\end{table}

\subsubsection{Ablation Study}

\begin{table}[t]
    \centering
    \caption{Impact of the pairwise loss term on the performance of PSH for document hashing in terms of precision.}\vspace{-0.05in}
    \resizebox{\columnwidth}{!}{%
    \begin{tabular}{c||c|c|c|c|c}
    \hline
         Loss weight ($\beta$) & 0 & 0.05 & 0.075 & 0.085 & 0.09  \\
         \hline\hline
         Precision & 0.8280  &  0.8373 &  0.7925  &  0.7340 & 0.7154 \\
         \hline
    \end{tabular}%
    }
    \label{tab:pw}
\end{table}

\begin{table}[h!]
    \centering
    \caption{Impact of the KL term on the performance of PSH for document hashing in terms of precision.}\vspace{-0.05in}
    \resizebox{\columnwidth}{!}{%
    \begin{tabular}{c||c|c|c|c|c|c}
    \hline
         KL weight ($\lambda$) & 0 & 0.01 & 0.1 & 0.5 & 1 & 2  \\
         \hline\hline
         Precision & 0.7712 & 0.8376  & 0.4954  &  0.4474  & 0.4870  & 0.4167 \\
         \hline
    \end{tabular}%
    }
    \label{tab:kl}
\end{table}

In this section, we perform ablation studies on the impacts of the pairwise loss and KL regularization terms on the performance of PSH with 32-bit hashing code. Table~\ref{tab:pw}
shows the precision of PSH for document retrieval on the 20 Newsgroup dataset for various pairwise loss weight $\beta$ values. We observe that discarding the pairwise loss term ($\beta=0$) decreases the performance of the PSH in learning effective hashing codes for document retrieval. Similarly, increasing $\beta$ to values higher than 0.05 degrades the performance significantly, indicating the importance of cross-validating the weight of the pairwise loss term. 

Table~\ref{tab:kl} illustrates the performance of PSH for document retrieval on the 20 Newsgroup dataset for various KL regularization weight $\lambda$ values, indicating the sensitivity of PSH to the weight of the KL regularization term. Specifically, PSH achieves the best performance for small $\lambda$ values. This observation is consistent with the literature \citep{zhao2017infovae,alemi2018fixing}, where KL weights less than one are associated with maximizing the mutual information between the observations and latent variables, hence increasing the effectiveness of hashing codes. 

\subsubsection{Qualitative Analysis of Semantic Information}
Similar to \citet{shen2018nash} and \citet{miao2016nvdm}, we examine the nearest neighbors of some words in the word vector space learned on 20 Newsgroup dataset. We calculate the distances based on the (word embedding) matrix $E \in \R^{K \times |V|}$ and select top 4 of the nearest neighbors. The results for ARM-DVAE and NASH are provided in Table \ref{tab:semantic}. We can see that our method places semantically-similar words closer together in the embedding space.

\begin{table}[t]
    \centering
    \caption{The four nearest neighbors in the word embedding space.}\vspace{-0.05in}
    \resizebox{\columnwidth}{!}{%
    \begin{tabular}{c||c|c|c|c}
    \hline
         Method/Word & weapons & medical & companies & book \\
         \hline\hline
          & guns & treatment & market & books \\
          & weapon & therapy & company & letters \\
         ARM-DVAE & violent & medicine & customers & references \\
          & rifles & hospital & industry & subject \\
         \hline
          & gun & treatment & company & books \\
          & guns &disease & market & english\\
         NASH & weapon & drugs & afford & references\\
          & armed & health & products & learning\\
    \hline
    \end{tabular}
    }
    \label{tab:semantic}
\end{table}

\subsection{Computational Complexity}
Our proposed framework for both supervised (PSH-ARM) and unsupervised (ARM-DVAE) semantic hashing can effectively be applied to large-scale datasets. To demonstrate this property, we apply both models on a collection of documents from the RCV1 benchmark \citep{lewis2004rcv1} with 100,000 training documents and 20,000 test documents. Table~\ref{tab:rcv} includes the precision at 100 of PSH-ARM and ARM-DVAE on the RCV1 dataset for various hashing code lengths. Both methods achieve high precision values for different hashing lengths, with PSH-ARM achieving close to 0.98, indicating the effectiveness of our framework. The run-time of each epoch in the training phase for PSH-ARM and ARM-DVAE is around 0.6 and 2 minutes, respectively.

\begin{table}[t]
    \centering
    \caption{The performances of the proposed unsupervised (ARM-DVAE) and supervised (PSH-ARM) hashing models on RCV1 dataset with different hashing code lengths.}\vspace{-0.05in}
    \resizebox{\columnwidth}{!}{%
    \begin{tabular}{c||c|c|c|c|c}
    \hline
         Method & 8 bits & 16 bits & 32 bits & 64 bits & 128 bits  \\
         \hline\hline
         PSH-ARM &  0.9754 & 0.9788  & 0.9782  & 0.9737  & 0.9759 \\
         \hline
         ARM-DVAE & 0.8368 & 0.8899  & 0.8988  & 0.8993 & 0.8968 \\
         \hline\hline
    
    \hline
    \end{tabular}%
    }
    \label{tab:rcv}
\end{table}

 \section{CONCLUSION}
In this paper, we exploit Augment-REINFORCE-Merge (ARM), an unbiased, low-variance gradient estimator to build effective semantic hashing with a discrete latent VAE. Employing the ARM gradient leads to not only fast convergence, but also low negative evidence lower bounds for variational inference, thus increasing the ability to reconstruct the original word counts from the latent hashing codes. More critically, we propose PSH by adding a pairwise loss function to the base discrete VAE to reward within-class similarity and between-class dissimilarity in the supervised hashing setting.  We conduct comprehensive experiments on several benchmark datasets, including the large-scale RCV1 benchmark, for both supervised and unsupervised hashing and show the superior performance of our proposed model in terms of precision for document retrieval.

\bibliographystyle{plainnat}
\bibliography{references,reference2}

\begin{thebibliography}{33}
\providecommand{\natexlab}[1]{#1}
\providecommand{\url}[1]{\texttt{#1}}
\expandafter\ifx\csname urlstyle\endcsname\relax
  \providecommand{\doi}[1]{doi: #1}\else
  \providecommand{\doi}{doi: \begingroup \urlstyle{rm}\Url}\fi

\bibitem[Alemi et~al.(2018)Alemi, Poole, Fischer, Dillon, Saurous, and
  Murphy]{alemi2018fixing}
Alexander Alemi, Ben Poole, Ian Fischer, Joshua Dillon, Rif~A Saurous, and
  Kevin Murphy.
\newblock Fixing a broken {ELBO}.
\newblock In \emph{International Conference on Machine Learning}, pages
  159--168, 2018.

\bibitem[Andoni(2009)]{andoni2009nearest}
Alexandr Andoni.
\newblock \emph{Nearest neighbor search: the old, the new, and the impossible}.
\newblock PhD thesis, Massachusetts Institute of Technology, 2009.

\bibitem[Bengio et~al.(2013)Bengio, L{\'e}onard, and
  Courville]{bengio2013estimating}
Yoshua Bengio, Nicholas L{\'e}onard, and Aaron Courville.
\newblock Estimating or propagating gradients through stochastic neurons for
  conditional computation.
\newblock \emph{arXiv preprint arXiv:1308.3432}, 2013.

\bibitem[Boluki et~al.(2020)Boluki, Ardywibowo, Dadaneh, Zhou, and
  Qian]{boluki2020learnable}
Shahin Boluki, Randy Ardywibowo, Siamak~Zamani Dadaneh, Mingyuan Zhou, and
  Xiaoning Qian.
\newblock Learnable {B}ernoulli dropout for {B}ayesian deep learning.
\newblock \emph{arXiv preprint arXiv:2002.05155}, 2020.

\bibitem[Chaidaroon and Fang(2017)]{chaidaroon2017variational}
Suthee Chaidaroon and Yi~Fang.
\newblock Variational deep semantic hashing for text documents.
\newblock In \emph{Proceedings of the 40th International ACM SIGIR Conference
  on Research and Development in Information Retrieval}, pages 75--84. ACM,
  2017.

\bibitem[Dadaneh et~al.(2020)Dadaneh, Boluki, Zhou, and Qian]{icassp_arsm}
Siamak~Zamani Dadaneh, Shahin Boluki, Mingyuan Zhou, and Xiaoning Qian.
\newblock Arsm gradient estimator for supervised learning to rank.
\newblock In \emph{ICASSP 2020 - 2020 IEEE International Conference on
  Acoustics, Speech and Signal Processing (ICASSP)}, pages 3157--3161, 2020.

\bibitem[Datar et~al.(2004)Datar, Immorlica, Indyk, and
  Mirrokni]{datar2004locality}
Mayur Datar, Nicole Immorlica, Piotr Indyk, and Vahab~S Mirrokni.
\newblock Locality-sensitive hashing scheme based on p-stable distributions.
\newblock In \emph{Proceedings of the twentieth Annual Symposium on
  Computational Geometry}, pages 253--262. ACM, 2004.

\bibitem[Dieng et~al.(2019)Dieng, Kim, Rush, and Blei]{dieng2019avoiding}
Adji~B Dieng, Yoon Kim, Alexander~M Rush, and David~M Blei.
\newblock Avoiding latent variable collapse with generative skip models.
\newblock In \emph{The 22nd International Conference on Artificial Intelligence
  and Statistics}, pages 2397--2405, 2019.

\bibitem[Higgins et~al.(2017)Higgins, Matthey, Pal, Burgess, Glorot, Botvinick,
  Mohamed, and Lerchner]{higgins2017beta}
Irina Higgins, Loic Matthey, Arka Pal, Christopher Burgess, Xavier Glorot,
  Matthew Botvinick, Shakir Mohamed, and Alexander Lerchner.
\newblock beta-{VAE}: Learning basic visual concepts with a constrained
  variational framework.
\newblock \emph{ICLR}, 2\penalty0 (5):\penalty0 6, 2017.

\bibitem[Jang et~al.(2016)Jang, Gu, and Poole]{jang2016categorical}
Eric Jang, Shixiang Gu, and Ben Poole.
\newblock Categorical reparameterization with {G}umbel-softmax.
\newblock \emph{arXiv preprint arXiv:1611.01144}, 2016.

\bibitem[Kingma and Ba(2014)]{kingma2014adam}
Diederik~P Kingma and Jimmy Ba.
\newblock Adam: A method for stochastic optimization.
\newblock \emph{arXiv preprint arXiv:1412.6980}, 2014.

\bibitem[Kingma and Welling(2013)]{kingma2013auto}
Diederik~P Kingma and Max Welling.
\newblock Auto-encoding variational {B}ayes.
\newblock \emph{arXiv preprint arXiv:1312.6114}, 2013.

\bibitem[LeCun et~al.(2015)LeCun, Bengio, and Hinton]{lecun2015deep}
Yann LeCun, Yoshua Bengio, and Geoffrey Hinton.
\newblock Deep learning.
\newblock \emph{Nature}, 521\penalty0 (7553):\penalty0 436, 2015.

\bibitem[Lew et~al.(2006)Lew, Sebe, Djeraba, and Jain]{lew2006content}
Michael~S Lew, Nicu Sebe, Chabane Djeraba, and Ramesh Jain.
\newblock Content-based multimedia information retrieval: State of the art and
  challenges.
\newblock \emph{ACM Transactions on Multimedia Computing, Communications, and
  Applications (TOMM)}, 2\penalty0 (1):\penalty0 1--19, 2006.

\bibitem[Lewis et~al.(2004)Lewis, Yang, Rose, and Li]{lewis2004rcv1}
David~D Lewis, Yiming Yang, Tony~G Rose, and Fan Li.
\newblock Rcv1: A new benchmark collection for text categorization research.
\newblock \emph{Journal of machine learning research}, 5\penalty0
  (Apr):\penalty0 361--397, 2004.

\bibitem[Liu et~al.(2011)Liu, Wang, Kumar, and Chang]{liu2011hashing}
Wei Liu, Jun Wang, Sanjiv Kumar, and Shih-Fu Chang.
\newblock Hashing with graphs.
\newblock In \emph{Proceedings of the 28th International Conference on
  International Conference on Machine Learning}, pages 1--8. Omnipress, 2011.

\bibitem[Liu et~al.(2012)Liu, Wang, Ji, Jiang, and Chang]{liu2012supervised}
Wei Liu, Jun Wang, Rongrong Ji, Yu-Gang Jiang, and Shih-Fu Chang.
\newblock Supervised hashing with kernels.
\newblock In \emph{2012 IEEE Conference on Computer Vision and Pattern
  Recognition}, pages 2074--2081. IEEE, 2012.

\bibitem[Maddison et~al.(2016)Maddison, Mnih, and Teh]{maddison2016concrete}
Chris~J Maddison, Andriy Mnih, and Yee~Whye Teh.
\newblock The concrete distribution: A continuous relaxation of discrete random
  variables.
\newblock \emph{arXiv preprint arXiv:1611.00712}, 2016.

\bibitem[McInnes et~al.(2018)McInnes, Healy, and Melville]{mcinnes2018umap}
Leland McInnes, John Healy, and James Melville.
\newblock Umap: Uniform manifold approximation and projection for dimension
  reduction.
\newblock \emph{arXiv preprint arXiv:1802.03426}, 2018.

\bibitem[Miao et~al.(2016)Miao, Yu, and Blunsom]{miao2016nvdm}
Yishu Miao, Lei Yu, and Phil Blunsom.
\newblock Neural variational inference for text processing.
\newblock In \emph{Proceedings of the 33rd International Conference on
  International Conference on Machine Learning - Volume 48}, ICML’16, page
  1727–1736. JMLR.org, 2016.

\bibitem[Pandey et~al.(2009)Pandey, Broder, Chierichetti, Josifovski, Kumar,
  and Vassilvitskii]{pandey2009nearest}
Sandeep Pandey, Andrei Broder, Flavio Chierichetti, Vanja Josifovski, Ravi
  Kumar, and Sergei Vassilvitskii.
\newblock Nearest-neighbor caching for content-match applications.
\newblock In \emph{Proceedings of the 18th International Conference on World
  Wide Web (WWW)}, pages 441--450. ACM, 2009.

\bibitem[Ramos et~al.(2003)]{ramos2003using}
Juan Ramos et~al.
\newblock Using tf-idf to determine word relevance in document queries.
\newblock In \emph{Proceedings of the first instructional conference on machine
  learning}, volume 242, pages 133--142. Piscataway, NJ, 2003.

\bibitem[Rezende et~al.(2014)Rezende, Mohamed, and
  Wierstra]{rezende2014stochastic}
Danilo~Jimenez Rezende, Shakir Mohamed, and Daan Wierstra.
\newblock Stochastic backpropagation and approximate inference in deep
  generative models.
\newblock In \emph{International Conference on Machine Learning}, pages
  1278--1286, 2014.

\bibitem[Salakhutdinov and Hinton(2009)]{salakhutdinov2009semantic}
Ruslan Salakhutdinov and Geoffrey Hinton.
\newblock Semantic hashing.
\newblock \emph{International Journal of Approximate Reasoning}, 50\penalty0
  (7):\penalty0 969--978, 2009.

\bibitem[Sarwar et~al.(2001)Sarwar, Karypis, Konstan, Riedl,
  et~al.]{sarwar2001item}
Badrul~Munir Sarwar, George Karypis, Joseph~A Konstan, John Riedl, et~al.
\newblock Item-based collaborative filtering recommendation algorithms.
\newblock In \emph{Proceedings of the 10th International Conference on World
  Wide Web (WWW)}, pages 285--295, 2001.

\bibitem[Shen et~al.(2018)Shen, Su, Chapfuwa, Wang, Wang, Henao, and
  Carin]{shen2018nash}
Dinghan Shen, Qinliang Su, Paidamoyo Chapfuwa, Wenlin Wang, Guoyin Wang,
  Ricardo Henao, and Lawrence Carin.
\newblock {NASH: T}oward end-to-end neural architecture for generative semantic
  hashing.
\newblock In \emph{Proceedings of the 56th Annual Meeting of the Association
  for Computational Linguistics (Volume 1: Long Papers)}, pages 2041--2050,
  2018.

\bibitem[Srivastava et~al.(2014)Srivastava, Hinton, Krizhevsky, Sutskever, and
  Salakhutdinov]{srivastava2014dropout}
Nitish Srivastava, Geoffrey Hinton, Alex Krizhevsky, Ilya Sutskever, and Ruslan
  Salakhutdinov.
\newblock Dropout: {A} simple way to prevent neural networks from overfitting.
\newblock \emph{The Journal of Machine Learning Research}, 15\penalty0
  (1):\penalty0 1929--1958, 2014.

\bibitem[Wang et~al.(2010)Wang, Kumar, and Chang]{wang2010semi}
Jun Wang, Sanjiv Kumar, and Shih-Fu Chang.
\newblock Semi-supervised hashing for scalable image retrieval.
\newblock In \emph{2010 IEEE Computer Society Conference on Computer Vision and
  Pattern Recognition}, pages 3424--3431. IEEE, 2010.

\bibitem[Wang et~al.(2013)Wang, Zhang, and Si]{wang2013semantic}
Qifan Wang, Dan Zhang, and Luo Si.
\newblock Semantic hashing using tags and topic modeling.
\newblock In \emph{Proceedings of the 36th International ACM SIGIR Conference
  on Research and Development in Information Retrieval}, pages 213--222. ACM,
  2013.

\bibitem[Weiss et~al.(2009)Weiss, Torralba, and Fergus]{weiss2009spectral}
Yair Weiss, Antonio Torralba, and Rob Fergus.
\newblock Spectral hashing.
\newblock In \emph{Advances in Neural Information Processing Systems}, pages
  1753--1760, 2009.

\bibitem[Yin and Zhou(2019)]{yin2019arm}
Mingzhang Yin and Mingyuan Zhou.
\newblock {ARM: Augment-REINFORCE-merge} gradient for stochastic binary
  networks.
\newblock In \emph{International Conference on Learning Representations}, 2019.

\bibitem[Zhang et~al.(2010)Zhang, Wang, Cai, and Lu]{zhang2010self}
Dell Zhang, Jun Wang, Deng Cai, and Jinsong Lu.
\newblock Self-taught hashing for fast similarity search.
\newblock In \emph{Proceedings of the 33rd International ACM SIGIR Conference
  on Research and Development in Information Retrieval}, pages 18--25. ACM,
  2010.

\bibitem[Zhao et~al.(2017)Zhao, Song, and Ermon]{zhao2017infovae}
Shengjia Zhao, Jiaming Song, and Stefano Ermon.
\newblock Infovae: Information maximizing variational autoencoders.
\newblock \emph{arXiv preprint arXiv:1706.02262}, 2017.

\end{thebibliography}

\end{document}